\documentclass[journal]{IEEEtran}
\usepackage{amsmath,amsfonts,amssymb}
\usepackage{algorithmic}
\usepackage{algorithm}
\usepackage{array}
\usepackage[caption=false,font=normalsize,labelfont=sf,textfont=sf]{subfig}
\usepackage{textcomp}
\usepackage{stfloats}
\usepackage{url}
\usepackage{verbatim}
\usepackage{graphicx}
\usepackage{cite}
\usepackage{xcolor}
\usepackage[utf8]{inputenc}
\usepackage{booktabs}
\usepackage[T1]{fontenc}
\usepackage{amsmath,amsfonts,amsthm} 
\usepackage{xcolor}

\usepackage[normalem]{ulem} % only for \sout in \zsedit below

\begin{document}

\title{Scalable Rao-Blackwellized Online Planning for \\ High-Dimensional POMDPs}

\author{\fontsize{11pt}{18pt}\selectfont Jiho Lee$^{1}$, Nisar Ahmed$^{1}$, Kyle Hollins Wray$^{2}$, Zachary Sunberg$^{1}$% <-this % stops a space
\thanks{$^{1}$Ann And H.J. Smead Aerospace Engineering Sciences Department, University of Colorado Boulder, Boulder, CO, USA. {\tt\small Email: \{Jiho.Lee, Nisar.Ahmed, Zachary.Sunberg\}@colorado.edu}}
\thanks{$^{2}$Manning College of Information and Computer Sciences, University of Massachusetts Amherst, Amherst, MA, USA. {\tt\small Email: kwray@umass.edu}}}
\maketitle
        % <-this % stops a space
%\thanks{}% <-this % stops a space
%\thanks{Manuscript received April 19, 2021; revised August 16, 2021.}}

% \markboth{IEEE TRANSACTIONS ON ROBOTICS}%


\maketitle

\begin{abstract}
Online planning under uncertainty remains a fundamental challenge for robotic systems operating in partially observable environments with high-dimensional state spaces. While sampling-based POMDP solvers enable approximate decision-making in large or continuous domains, their performance degrades as belief dimensionality increases due to the high variance inherent in Monte Carlo-based estimation. In this work, we extend the Rao-Blackwellized online POMDP (RB-POMDP) framework to improve its generalizability in high-dimensional settings through hybrid continuous–discrete belief representations. By analytically propagating uncertainty associated with marginalized state components during tree-based planning, the proposed approach reduces sampling-induced variance in value estimation. We demonstrate the effectiveness of this framework in a robotic search-and-rescue task by integrating it with FastSLAM~2.0. Experimental results show that the proposed planner achieves higher cumulative rewards using significantly fewer particles and planning simulations than purely sampling-based methods under equivalent computational budgets. These results suggest that structured high-dimensional robotic problems admitting tractable sufficient statistics can be effectively leveraged within the RB-POMDP framework for computationally feasible online decision-making.
\end{abstract}


\section{Introduction}
\IEEEPARstart{R}{obotic} decision-making tasks such as search-and-rescue, autonomous navigation, and exploration often require reasoning under significant uncertainty arising from limited sensing and incomplete knowledge of the environment\cite{agha-mohammadi_slap_2018}\cite{meera_obstacle-aware_2019}\cite{flaspohler_information-guided_2019}\cite{placed_survey_2023}\cite{zhitnikov_simplified_2024}. Partially Observable Markov Decision Processes (POMDPs) provide a powerful mathematical framework for sequential decision-making under uncertainty by enabling agents to maintain beliefs over latent system states while planning actions that maximize expected long-term reward \cite{kochenderfer_algorithms_2022}. While POMDPs have been widely applied to various domains \cite{sunberg_value_2017}\cite{durrant-whyte_unmanned_2012}\cite{wray_online_2017}, their practical deployment still remains challenging due to the computational burden associated with belief representation and online planning in large-scale environments.

Sampling-based online POMDP solvers such as partially observable Monte Carlo planning (POMCP) \cite{silver_monte-carlo_2010} and its variant partially observable Monte Carlo planning with observation widening (POMCPOW)\cite{sunberg_online_2018} maintain particle-based belief representations and rely on Monte Carlo simulation to approximate value estimates during online tree search. Recent theoretical results show that these planners can achieve convergence guarantees that depend only on the number of samples and are independent of state or observation space dimensionality, provided that the importance weight is bounded \cite{lim_optimality_2023}. In practice, however, maintaining bounded importance weights become extremely challenging in high-dimensional settings, where probability mass concentrates in a small typical set of the state space and importance weights may vary significantly across samples \cite{jordan_introduction_1998}. As a result, particle filters often require a substantial increase in the number of particles to maintain reliable belief representation due to likelihood concentration and importance weight degeneracy \cite{snyder_obstacles_2008}\cite{surace_how_2019}\cite{bickel_sharp_2008-1}. In such settings, the effective sample size of the particles decreases, increasing the variance of estimators computed using the particles \cite{liu_monte_2008}. Since value estimates are computed via Monte Carlo runs under the particle belief, a low effective sample size increases the variance in value estimates in tree search. Consequently, consistent long-horizon planning requires a substantial increase in both the number of particles used for belief representation and the number of tree simulations needed for reliable planning. This leads to significant computational overhead and often renders these purely sampling-based approaches impractical for realistic robotic decision-making scenarios.

\begin{figure}[t]
    \centering
    \includegraphics[width=\linewidth]{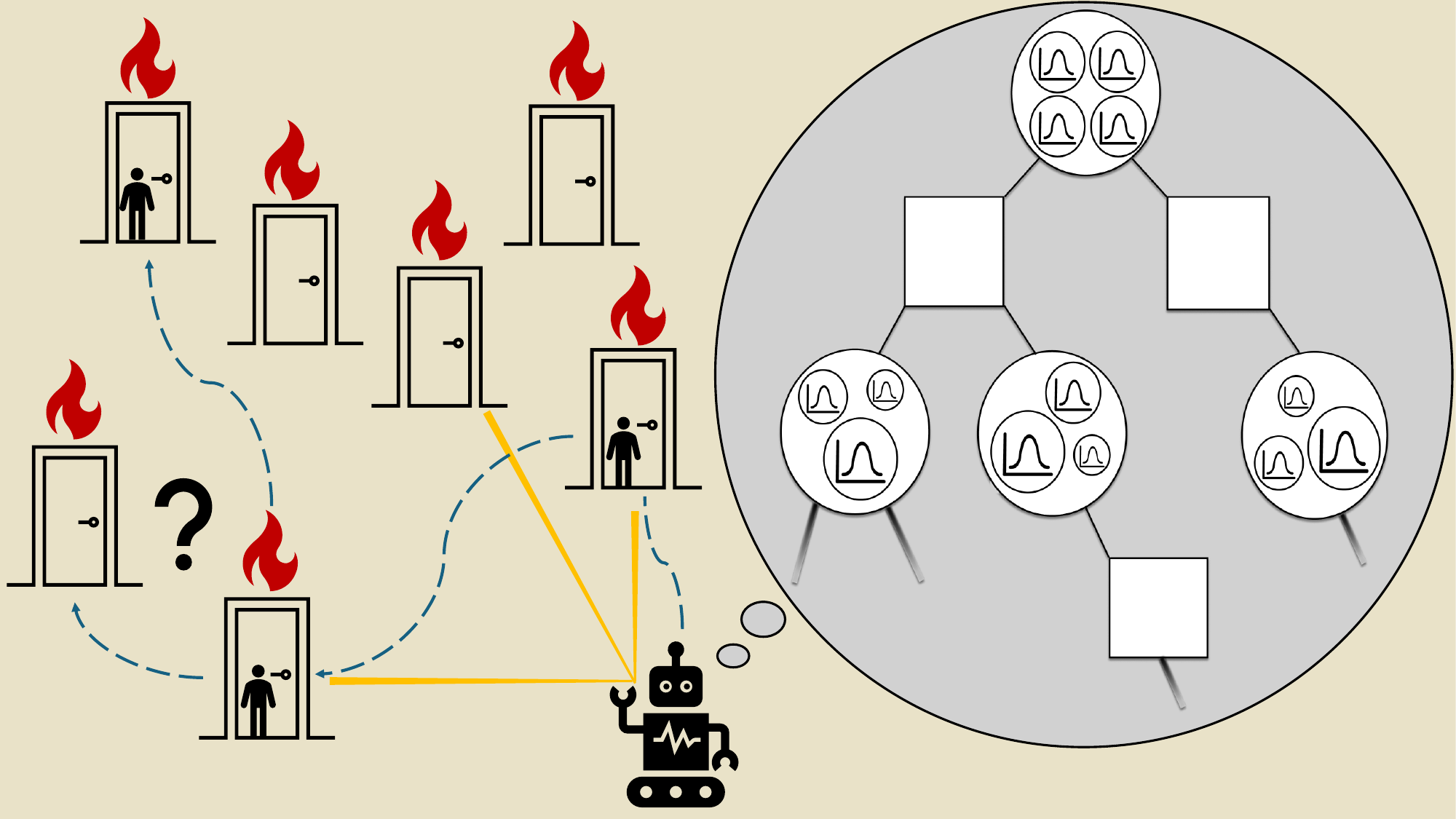}
    \caption{Motivating scenario for search-and-rescue under partial observability. A robot navigates an indoor environment consisting of multiple rooms while constructing a map and localizing itself. Victim presence behind each door is unknown and can only be inferred through limited sensing of both spatial and victim-related semantic cues. To accomplish the task, the agent maintains a belief over the joint robot and environment state and considers multiple possible future outcomes under different action-observation histories, represented by the planning tree on the right, to select actions that locate victims efficiently.}
    \label{fig:experimental_setup}
\end{figure}

Many robotic systems, however, exhibit structured latent state spaces in which subsets of variables admit conditionally tractable posterior updates. For example, problems such as simultaneous localization and mapping (SLAM) involve latent geometric variables (e.g., landmark positions) whose uncertainty can be represented analytically given a sampled robot trajectory \cite{montemerlo_fastslam_2003}. Such conditional structure implies that portions of the state admit sufficient statistics that can be propagated in closed form, rather than requiring full Monte Carlo approximation. Rao-Blackwellized Particle Filtering (RBPF) exploits this structure by analytically marginalizing subsets of the latent state conditioned on sampled variables, reducing the dimensionality that must be explored via particles \cite{harvey_sequential_2005}\cite{doucet_rao-blackwellised_2001}.

RBPF has been successfully applied to various estimation problems such as SLAM and target tracking \cite{montemerlo_fastslam_2002}\cite{sarkka_rao-blackwellized_2007}\cite{stachniss_information_2005}. Within the context of POMDPs, a recent work \cite{lee_rao-blackwellized_2025} introduced a Rao-Blackwellized POMDP (RB-POMDP) online planning framework that leverages analytical belief updates within the POMCPOW planner. While this approach demonstrated improved efficiency in continuous state and observation planning tasks, it remained limited to Kalman filter-based analytical components and relatively simple, low-dimensional experimental settings. Therefore, it is of significant interest to explore how this framework can provide a general mechanism to enable scalable decision-making in realistic robotic problems with much larger latent state spaces and non-Gaussian distributions. 

This paper addresses this gap by extending the RB-POMDP framework to support arbitrary analytically tractable belief components and demonstrating the feasibility of online planning in high-dimensional problems. By analytically propagating uncertainty associated with marginalized state variables during tree-based planning, the new framework reduces sampling-induced variance in reward estimates, enabling consistent long-horizon reasoning with substantially fewer particles and tree iterations. 
In summary, this work makes the following three contributions: 
\begin{itemize}
    \item[1)] We demonstrate that Rao-Blackwellized online planning is not merely beneficial but essential for computationally feasible decision-making in high-dimensional POMDPs with analytically tractable structure. Scalable planning becomes possible where purely sampling-based methods are infeasible under practical computational budgets. 
    \item [2)] We generalize the RB-POMDP framework beyond Gaussian assumptions by allowing the analytically maintained component to be instantiated with arbitrary closed-form belief update models. This enables planning over hybrid continuous-discrete latent state spaces in which marginalized variables may follow non-Gaussian distributions.
    \item [3)] We validate the proposed framework in a realistic search-and-rescue scenario involving both geometric and semantic observations by integrating it with the FastSLAM 2.0 algorithm, as illustrated in Figure \ref{fig:experimental_setup}. This scenario induces a hybrid belief over robot pose, landmark geometry, and semantic victim presence, represented by particles, Gaussian, and Bernoulli distributions, respectively. This demonstrates that any structured high-dimensional estimation problems, if admitting tractable sufficient statistics, can be leveraged within the RB-POMDP framework. 
\end{itemize}

The remainder of this article is organized as follows. Section~\ref{background} reviews POMDPs, sampling-based POMDP planning methods, and RBPFs. Section~\ref{technical_approach} introduces the Rao-Balckwellized factorization of POMDPs, RBPF-based belief updates, and Rao-Blackwellized online planning. Section~\ref{problem_formulation} presents the search-and-rescue problem application, followed by Section~\ref{Methodology}, which details the proposed Rao-Blackwellized belief representation and associated planning components. Experimental results are provided in Section~\ref{experimental_results}. Finally, Section~\ref{conclusion} discusses the results and outlines directions for future work.

\section{Background}
\label{background}
\subsection{Partially Observable Markov Decision Process (POMDP)}
In Markov Decision Processes (MDPs), agents operate with full knowledge of the current system state. At any given time, the agent fully understands where it is or what the situation is. In POMDPs, however, agents do not have perfect knowledge of the current state. They must make decisions under uncertainty about the current as well as future states; thus, POMDPs are more suitable for real-world robotics scenarios where the true state of the environment cannot be directly observed due to sensing limitations or environmental uncertainty \cite{thrun_probabilistic_2005}.

A POMDP is formally defined by the tuple $(\mathcal{S}, \mathcal{A}, \mathcal{T}, \mathcal{O}, \mathcal{R}, \mathcal{Z}, \gamma)$ where $\mathcal{S}$ is the state space, $\mathcal{A}$ is the action space, $\mathcal{T}$ is the state transition function, $\mathcal{O}$ is the observation space, $\mathcal{Z}$ is the observation function, $\mathcal{R}$ is the reward function, and $\gamma \in (0,1)$ is the discount factor. Because the agent cannot directly observe the system state, it instead maintains a probabilistic belief distribution over possible states. This belief is recursively updated using Bayes' rule:
\begin{equation}
b'(s') \propto P(o \mid s', a) \sum_{s \in \mathcal{S}} P(s' \mid s, a) b(s),
\end{equation}
where $b(s)$ and $b'(s')$ denote the prior and posterior beliefs before and after executing action $a$ and receiving observation $o$, respectively.

Solving POMDPs is therefore performed over belief states rather than fully observed states. The optimal value function, which guides the selection of the best action based on the current belief, is given by
\begin{equation}
\label{value_function}
V^*(b) = \max_{a \in \mathcal{A}} \left[  \mathcal{R}(b, a) + \gamma \sum_{o \in \mathcal{O}} P(o \mid b, a) V^*(b') \right],
\end{equation}
where $\mathcal{R}(b, a)$ is the expected immediate reward for taking action $a$ and $P(o | b, a)$ is the probability of receiving observation $o$ after taking action $a$. Consequently, the dimensionality and structure of the belief representation can directly influence the computational complexity of evaluating the value function during planning.

\subsection{Sampling-Based POMDP Online Planning}
Evaluating the optimal value function in \eqref{value_function} is computationally intractable for most real-world problems due to the high-dimensional nature of belief spaces and observations. Consequently, a large body of work has focused on online planning methods that compute approximate action values during execution by simulating future outcomes from the current belief state.

The well-known sampling-based solver POMCP \cite{silver_monte-carlo_2010} employs particle-filter-based belief representations with Monte Carlo tree search (MCTS) and relies on a black-box generative model to sample future states, observations, and rewards during tree expansion. Extensions such as POMCPOW incorporate progressive widening techniques to regulate tree growth and improve planning performance in continuous observation spaces \cite{sunberg_online_2018}. 

These sampling-based methods approximate value functions by averaging simulated rewards obtained from running $N$ tree search simulations to a depth $D$.
Specifically, the value at an observation-action history $h$ at depth $\tau$ is estimated using the set of simulations consistent with that history, $I(h)$, as
\begin{equation}
V(h) \approx \frac{1}{|I(h)|} \sum_{i\in I(h)} \sum_{d=\tau}^{D} \gamma^{d-\tau}  \mathcal{R}(s_{i,d}, a_{i,d}).
\end{equation}
However, Monte Carlo sampling exhibits a convergence rate of $\mathcal{O}(1/\sqrt{N})$, which can require a large number of simulations to achieve consistent estimates. Especially when operating over high-dimensional belief spaces, uncertainty in predicted observations and rewards during simulation can introduce substantial variance in estimated  values within the planning tree. As a result, significant increase in the number of tree simulations is needed to achieve consistent, reliable long-horizon planning, inevitably leading to considerable computational overhead. 

% \nra{note: not immediately clear from description of POMCP/POMCPOW that they are using particle filters -- described above as Monte Carlo, which could be mistaken by novice reviewer for `direct' Monte Carlo as opposed to SIS/sequential importance sampling. Probably worth making this connection in this section clearer? Distinction is key since reducing variance of importance weights is the secret sauce behind RBPF for online planning...}

\subsection{Rao-Blackwellized Particle Filter (RBPF)}
One approach to reducing the variance of purely sampling-based Monte Carlo estimators is to exploit analytically tractable structure in the latent state using the Rao-Blackwell theorem.
This theorem states that conditioning an estimator on sufficient statistics cannot increase its variance \cite{ristic_beyond_2004}, as expressed by
\begin{equation}
    \text{var}[\delta(s)] \geq \text{var}[\delta(s \mid \Theta)],
\end{equation}
where $\delta(s)$ is any kind of estimator of the random variable $s$, and $\Theta$ denotes sufficient statistics for $s$.

In the context of sequential Monte Carlo estimation, i.e., particle filtering via sequential importance sampling, RBPF leverages this principle by analytically marginalizing subsets of the latent state that admit closed-form conditional updates. By maintaining sufficient statistics for these tractable components within each particle, RBPF reduces the dimensionality of the belief representation that must be approximated through sampling. This enables more efficient belief updates by reducing the variance of the resulting particle importance weights \cite{liu_monte_2008}\cite{ doucet_rao-blackwellised_2001}. As a result, this leads to using fewer particles compared to purely sampling-based approaches which sample all variables whether they possess analytically tractable conditional posterior sufficient statistics or not. %, but also requires higher computational effort per particle}.

\section{Technical Approach}
\label{technical_approach}
\begin{algorithm}[!t]
\caption{Belief Updates for Rao-Blackwellized Particle Filter adapted from \cite{lee_rao-blackwellized_2025}}
\label{alg:belief_update}
\begin{algorithmic}[1]
\REQUIRE $(s_k^{\pi,i}, w_k^{i}, \Theta_k^{\alpha|\pi,i})_{i=1}^{N_s}$, $o_{k+1}, a_{k+1} $
\ENSURE$(s_{k+1}^{\pi,i}, w_{k+1}^{i}, \Theta_{k+1}^{\alpha|\pi,i})_{i=1}^{N_s}$
\FOR{$i = 1$ \TO $N_s$}
    \STATE \text{Draw} $s_{k+1}^{\pi,i} \sim p(s_{k+1}^{\pi,i} | s_k^{\pi,i}, a_{k+1})$
    \STATE $\Theta_{k+1}^{\alpha|\pi,i}, \Lambda_{k+1}^{\alpha|\pi,i} \gets \text{AnalytUpd}\left(\Theta_k^{\alpha|\pi,i}, s_{k+1}^{\pi,i}, o_{k+1},a_{k+1}\right)$
    \STATE $\tilde{w}_{k+1}^{i} \gets \text{Reweight}\left(\Theta_{k+1}^{\alpha|\pi,i}, \Lambda_{k+1}^{\alpha|\pi,i}, s_{k+1}^{\pi,i}, o_{k+1}\right)$
\ENDFOR

\STATE $\left\{ w_{k+1}^{i} \right\}_{i=1}^{N_s} = \text{Normalize}\left( \left\{ \tilde{w}_{k+1}^{i} \right\}_{i=1}^{N_s} \right)$
\STATE \text{Compute} $N_{\text{ess}} = \frac{1}{\sum_{s=1}^{N_s} (w_{k+1}^{i})^2}$
\IF {$N_{\text{ess}} < \tau$}
    \STATE $\left\{ s_{k+1}^{\pi,i}, w_{k+1}^i \right\}_{i=1}^{N_s} \gets \text{Resample} \left( \left\{ s_{k+1}^{\pi,i}, w_{k+1}^i \right\}_{i=1}^{N_s} \right)$
\ENDIF
\end{algorithmic}
\end{algorithm}

\begin{figure}[!t]
\centerline{\includegraphics[width=1.0\columnwidth]{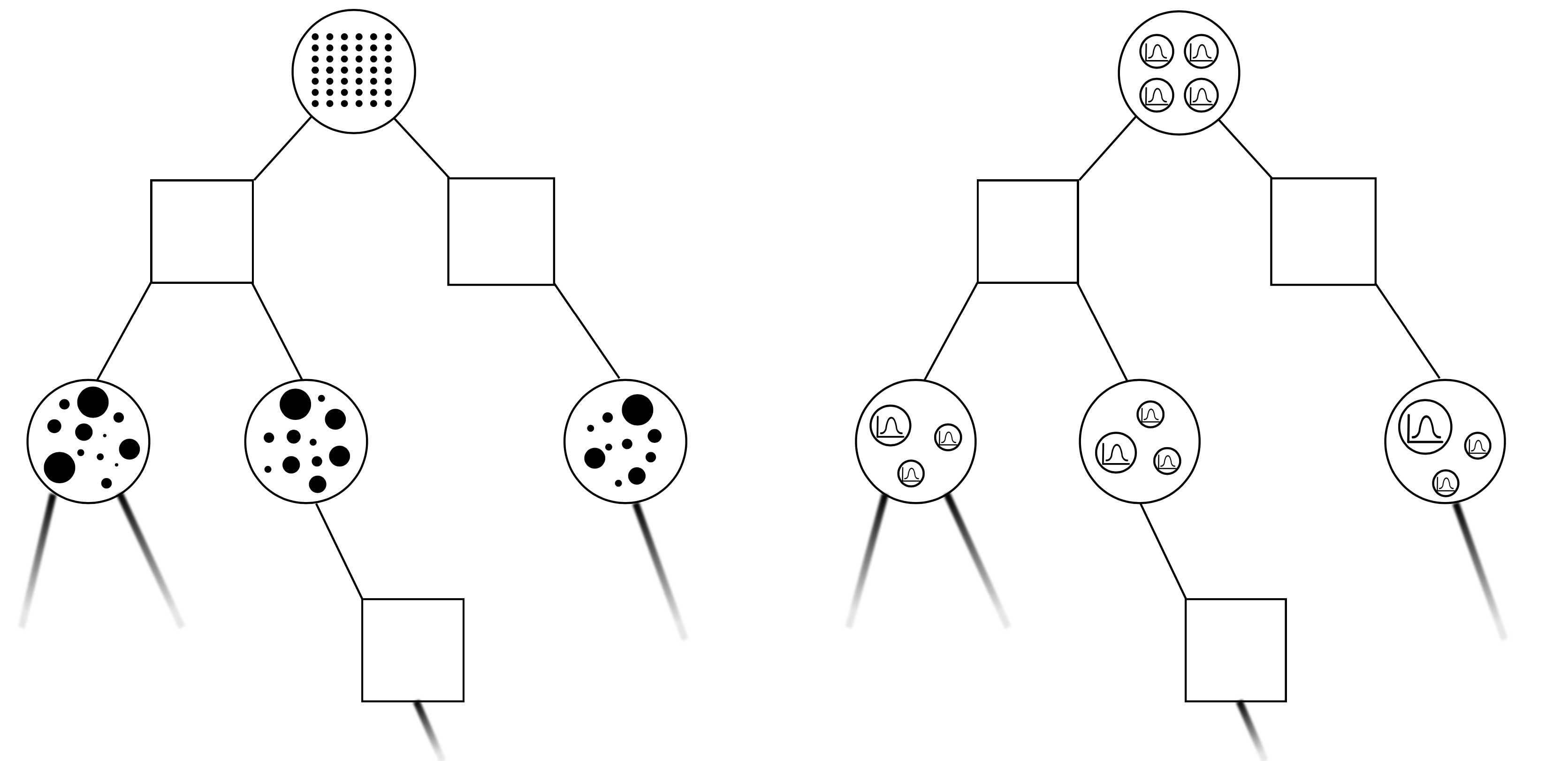}}
\caption{Tree structure comparison adapted from \cite{lee_rao-blackwellized_2025}. Each square and larger circle represents an action node and an observation node, respectively. In the POMCPOW tree, particles are shown as black dots while each particle in the RB-POMCPOW tree is associated with a Gaussian distribution. Due to the nature of POMCPOW, these particles form a weighted mixture of beliefs.}
\vspace{-0.4cm}
\label{fig:rbpf_tree_comparison}
\end{figure}

Incorporating analytically tractable belief components within sampling-based approximate POMDP solvers requires not only modifying how belief is represented but also altering how its conditional uncertainty is propagated during online planning. A recent work\cite{lee_rao-blackwellized_2025} introduced such an approach, referred to as the RB-POMDP framework, in which each particle is associated with a conditional analytical distribution over marginalized state variables as shown in Fig. \ref{fig:rbpf_tree_comparison}. This section describes the proposed method for addressing these challenges.

\subsection{Rao-Blackwell factorization of POMDPs}
The key idea underlying Rao-Blackwellization is to exploit conditional structure in the latent state by decomposing it into analytically tractable and non-tractable components. Let the state be partitioned as $s_t = (s_t^\pi, s_t^{\alpha|\pi})$, where $s_t^\pi$ denotes the subset of variables that must be represented through randomly sampled particles and $s_t^{\alpha|\pi}$ denotes variables whose conditional posterior admits a closed-form update. Using the chain rule, the joint posterior distribution over these components can be factorized as
\begin{equation}
\label{eq: factorization}
p(s_{t+1}^{\alpha|\pi}, s_{t+1}^\pi \mid o_{1:t}) =
p(s_{t+1}^{\alpha|\pi} \mid s_{t+1}^\pi, o_{1:t}) \,
p(s_{t+1}^\pi \mid o_{1:t}).
\end{equation}

This factorization enables analytical marginalization of the tractable component $s_{t+1}^{\alpha|\pi}$ conditioned on sampled realizations of $s_{t+1}^\pi$.
In other words, the conditional belief $p(s_{t+1}^{\alpha|\pi} \mid s_{t+1}^\pi, o_{1:t})$ can be propagated using closed-form updates (e.g., Kalman filtering). When such a decomposition exists, sampling is required only over the non-tractable component $s_{t+1}^\pi$, effectively reducing the dimensionality of the particles.

\subsection{RBPF Belief Updates}
In the most commonly used particle filtering variant, Sequential Importance Resampling particle filters (SIRPF) (also known as Bootstrap particle filters), a common choice for the proposal distribution is the transition function $p(s' \mid s, a)$, which leads to importance weights that are proportional to the observation likelihood \cite{doucet_introduction_2001}. In RBPFs, this likelihood is evaluated using the analytically updated belief over the tractable state components \cite{doucet_sequential_2001}, as indicated by $\Lambda$ in line 3 of Algorithm~\ref{alg:belief_update}. When using Kalman filters, for instance, this corresponds to the filter innovation likelihood function for a received observation and involves the innovation covariance matrix computed from the measurement update step.

When alternative proposal distributions are employed, such as the measurement-informed proposal used in FastSLAM 2.0, the importance weights must be computed with additional care. The corresponding computation for this likelihood is described in detail in Section~V for our application problem.

Unlike standard SIRPF approaches, where the weights are based solely on sampled observation likelihoods, the RBPF formulation leverages sufficient statistics ($\Theta$) maintained within each particle. This conditioning mechanism reduces the variance of importance weights \cite{liu_monte_2008} and thereby enables lower variance belief updates in accordance with the Rao-Blackwell theorem. The overall Rao-Blackwellized belief update procedure is summarized in Algorithm~\ref{alg:belief_update}.

\subsection{Online Planning with Rao-Blackwellization}

\begin{figure*}[t]
    \centering
    \includegraphics[width=\textwidth]{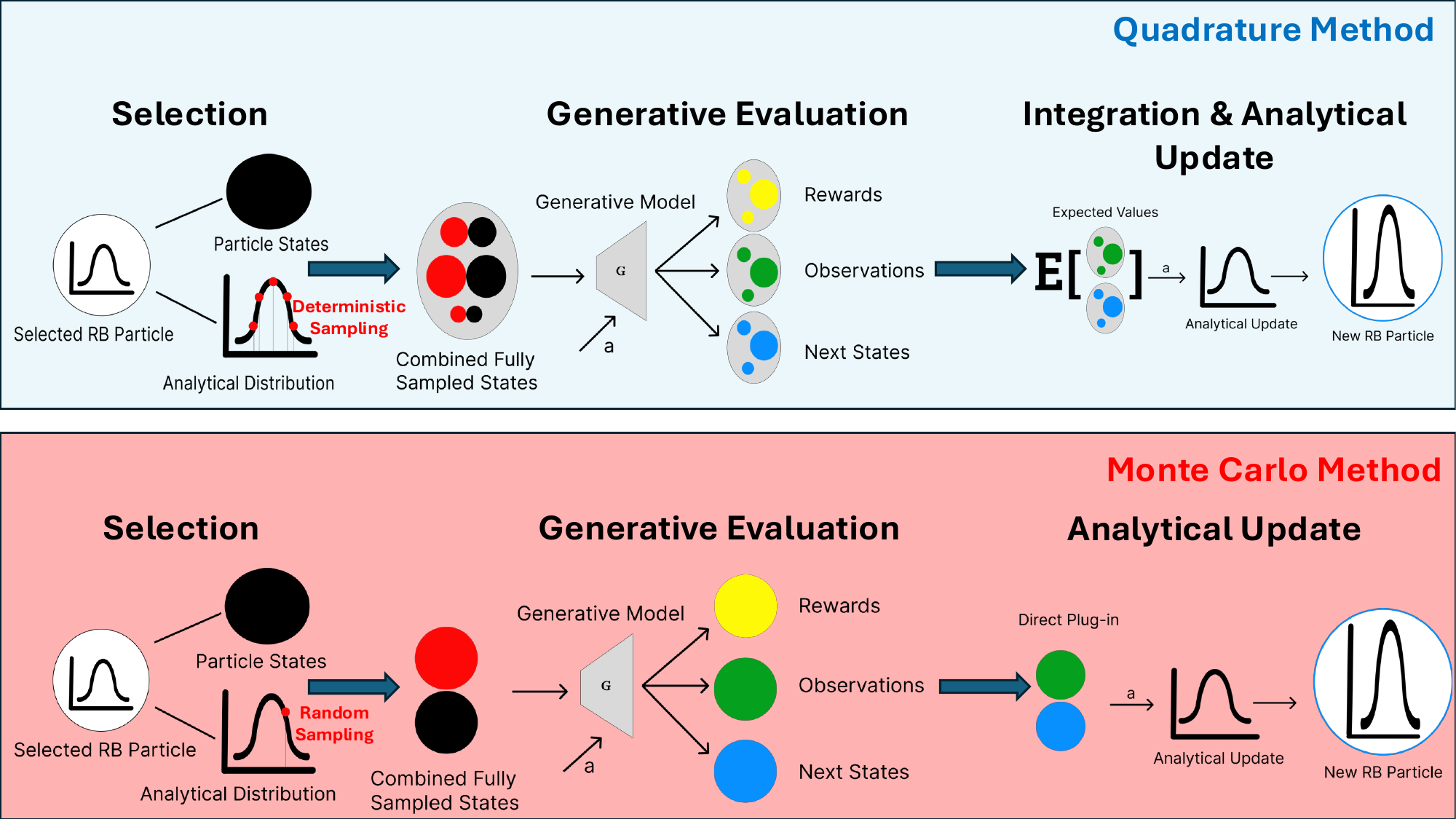}
    \caption{Illustration of the Rao-Blackwellized particle generative evaluation and analytical update during online planning. \textbf{Top}: Quadrature-based integration. A selected Rao-Blackwellized particle consists of sampled (particle-based) state components and analytically maintained conditional distributions. During generative evaluation, deterministic quadrature nodes are combined with the particle state to produce multiple fully instantiated and weighted next states, rewards, and observations. Although each realization yields a complete next state, only the sampled state components are used for the analytical belief update. The analytical state components are not averaged across these realizations; instead, they are updated using the expected observation and the expected next sampled component computed via their corresponding quadrature weights, yielding an updated analytical distribution. This results in a new Rao-Blackwellized particle stored at the corresponding tree node. \textbf{Bottom:} Monte Carlo evaluation. Analytically tractable components are randomly sampled to generate a single next-state realization. Integration is no longer needed, and the realized observation and next sampled state component is directly used for belief update. This introduces additional sampling-induced variance into predicted rewards, leading to slower convergence of value estimates and requiring substantially more tree simulations, as validated in Section \ref{planning_performance}.}
    \label{fig:rb_pipeline}
\end{figure*}

The key modification required to incorporate Rao-Blackwellized belief representations into sampling-based online POMDP solvers arises during generative evaluation inside the planning tree.
In standard POMCP and POMCPOW, rewards and observations are simulated by drawing a fully instantiated state realization from the particle-based belief at each tree expansion step.
However, when analytically tractable components of the state are maintained as conditional distributions within each Rao-Blackwellized particle, randomly sampling these variables from their distributions during simulation introduces additional uncertainty into predicted observations and rewards.

Instead, planning can be performed by computing the expected values with respect to the analytically maintained conditional distribution. For states consistent with a given history $h$, the immediate reward can be approximated as
\begin{equation}
      \mathcal{R}(h) \approx 
     \frac{1}{|I(h)|} \sum_{i \in I(h)} 
     \mathbb{E}_{s_{i}^{\alpha|\pi}}
     \left[ \mathcal{R}(s_{i}^{\pi}, s_{i}^{\alpha|\pi}, a_{i}) \right].
\end{equation}

A straightforward approach is to approximate this expectation via Monte Carlo sampling by drawing multiple realizations from the analytical distribution. However, drawing a large number of samples negates the computational benefits of Rao-Blackwellization, while again, using a single sample fails to adequately capture the underlying uncertainty and leads to high-variance value estimates. This increased variance degrades the accuracy of value estimates within the planning tree and typically requires substantially more tree simulations for convergence.

To address this issue, we instead approximate these expectations using deterministic quadrature methods. Quadrature techniques explicitly account for uncertainty in analytically tractable state components by replacing the expectation with a weighted sum over $M$ quadrature points $s_{i,d,k}^{\alpha|\pi}$ and corresponding weights $w_{i,d,k}$. The resulting value estimate is given by the following: 
\begin{equation}
    V(h) \approx \frac{1}{|I(h)|} 
    \sum_{i\in I(h)} 
    \sum_{d=\tau}^{D} 
    \gamma^{d-\tau} 
    \sum_{k=1}^{M} 
    w_{i,d,k} \,
    \mathcal{R}(s_{i,d}^{\pi}, s_{i,d,k}^{\alpha|\pi}, a_{i,d}).
\end{equation}
This comparison between quadrature and Monte Carlo methods is illustrated in Fig.~\ref{fig:rb_pipeline} which corresponds to lines 17 and 18 in $\textsc{Rollout}$ and lines 27 and 34 in $\textsc{Simulate}$ functions in Algorithm~\ref{alg:RB-POMCPOW}.

When using quadrature techniques, the choice of method depends on the distribution used for the analytically marginalized states. For example, we can employ Gaussian-Hermite quadrature to interpolate and perform the necessary integration when using Gaussian beliefs for the linear states\cite{liu_note_1994}. If different distributions were considered, other methods from the Askey family of orthogonal polynomials could be utilized \cite{szego_orthogonal_1939}\cite{schoutens_stochastic_2000}. 

To mitigate the curse of dimensionality associated with high-order quadrature, we employ Smolyak sparse grids, which reduce the number of quadrature points required while maintaining approximation accuracy \cite{barthelmann_high_2000}\cite{heiss_likelihood_2008}. The Smolyak formula for constructing the sparse grid is:
\begin{equation}
\mathcal{A}(q,d) = \sum_{q-d+1 \leq |\mathbf{i}| \leq q} (-1)^{q - |\mathbf{i}|} \binom{d-1}{q - |\mathbf{i}|} \bigotimes_{j=1}^{d} \mathcal{Q}_{i_j},
\end{equation} 
where $\mathcal{A}(q,d)$ is the set of Smolyak quadrature points and their corresponding weights for a given sparse grid level $q$ and dimensionality $d$, $i$ is a multi-index with $i_j$ representing the level of the univariate quadrature rules, and $ \bigotimes_{j=1}^{d} \mathcal{Q}_{i_j}$ denotes the tensor product of the univariate quadrature rules $\mathcal{Q}_{i_j}$ at levels $i_j$ for each dimension $j$.

Higher sparse grid levels yield more accurate expectation estimates at the cost of increased computational time. Hence, by choosing different levels of sparse grid, we can fine-tune the trade off between computational cost and integration accuracy in the planner. Overall, this approach reduces the reliance on Monte Carlo sampling for all states by capturing the marginalized states using deterministic quadrature techniques and ultimately reduces the total number of tree iterations needed in POMCP and POMCPOW.

The RB-POMCPOW pseudocode, adapted from \cite{lee_rao-blackwellized_2025}, is presented in Algorithm \ref{alg:RB-POMCPOW} with the modified code written in blue. Quadrature techniques are used in lines $17$ and $27$. Note that the sparse grid level or the integration methods can be adjusted independently for the $\textsc{Rollout}$ and $\textsc{Simulate}$ functions to compute the expectations. While this study primarily focuses on the RB-POMCPOW algorithm, the underlying principles for handling uncertainty in the analytical distributions of tractable states remain the same for POMCP (refer to Appendix \ref{RB-POMCP} for the pseudocode).

While quadrature techniques are used to approximate expectations for continuous analytically maintained belief components, the RB-POMDP framework is not restricted to such distributions. In particular, when marginalized variables have finite discrete support, expectations can be computed exactly. For example, if $X \sim \text{Bern}(p)$, then
\begin{equation}
\mathbb{E}[R(X)] = (1-p)R(0) + pR(1).
\end{equation}
This again helps avoid randomly sampling stochastic realizations when evaluating expected rewards and observations during tree expansion. In general, any marginalized state component admitting a closed-form conditional update can be incorporated within the proposed RB-POMDP framework.

\begin{algorithm*}[ht]
\caption{RB-POMCPOW adapted from \cite{lee_rao-blackwellized_2025}}
\label{alg:RB-POMCPOW}
{\small
\begin{algorithmic}[1]
\begin{minipage}[t]{0.48\textwidth}
\STATE \textbf{procedure} \textsc{Search}($h$)
    \STATE \quad \textbf{for} $i \leftarrow 1 \textbf{ to } N$ \textbf{do}
        \STATE \quad \quad \textcolor{blue}{\textsc{Simulate}($p \sim  b(h), h, d_{max}$)}
    \STATE \quad \textbf{return} $\arg\max_a Q(ha)$;
\STATE
\STATE \textbf{procedure} \textsc{ActionProgWiden}$(h)$
    \STATE \quad \textbf{if} $|\mathcal{C}(h)| \leq k_a N(h)^{\alpha_a}$ \textbf{then}
    \STATE \quad \quad $a \leftarrow \textsc{NextAction}(h)$
    \STATE \quad \quad $\mathcal{C}(h) \leftarrow \mathcal{C}(h) \cup \{a\}$
    \STATE \quad \textbf{return} ${\arg\max\limits}_{a \in \mathcal{C}(h)} \left[ Q(ha) + c\sqrt{\frac{\log N(h)}{N(ha)}} \right]$
\STATE
\STATE \textbf{procedure} \textsc{Rollout}($p, h, d$)
    \STATE \quad \textbf{if} $\gamma^{\text{depth}} < \epsilon$ \textbf{then}
        \STATE \quad \quad \textbf{return} 0
    \STATE \quad \textbf{end if}
    \STATE \quad $a \sim \pi_{\text{rollout}}(h, \cdot)$
    \STATE \quad \textcolor{blue}{$(\hat{s'}, \hat{o}, \hat{r}) \leftarrow \mathbb{E}[G(p, a)]$} \hfill \textcolor{black}{$\triangleright$ Quadrature Techniques} \phantom{xx}
    \STATE \quad \textcolor{blue}{$p' \leftarrow$ \textsc{AnalyticalUpdate}$(p, \hat{s'}, \hat{o}, a)$} 
    \STATE \quad \textcolor{blue}{\textbf{return} $\hat{r} + \gamma \cdot \textsc{Rollout}(p', ha\hat{o}, d - 1)$}
\STATE \textbf{end procedure}
\STATE 
\STATE \textbf{procedure} \textsc{Simulate}($p, h, d$)
    \STATE \quad \textbf{if} $d = 0$ \textbf{then}
        \STATE \quad \quad \textbf{return} 0
    \STATE \quad \textbf{end if}
\end{minipage}
\hfill
\begin{minipage}[t]{0.48\textwidth}
    \STATE \quad $a \leftarrow \textsc{ActionProgWiden}(h)$
    %\STATE \quad \textcolor{blue}{$\hat{s} \leftarrow \mathbb{E}[p]$} \hfill \textcolor{black}{$\triangleright$ Quadrature Techniques}
   \STATE \quad \textcolor{blue}{$(\hat{s'}, \hat{o}, \hat{r}) \leftarrow \mathbb{E}[G(p, a)]$} \hfill \textcolor{black}{$\triangleright$ Quadrature Techniques}
    \STATE \quad \textbf{if} $|C(ha)| \le k_o N(ha)^{\alpha_o}$ \textbf{then}
        \STATE \quad \quad \textcolor{blue}{$o \leftarrow \hat{o}$}
        \STATE \quad \quad $M(hao) \leftarrow M(hao) + 1$
    \STATE \quad \textbf{else}
        \STATE \quad \quad $o \leftarrow \text{select } o \in C(ha) \text{ w.p. } \frac{M(hao)}{\sum_{o} M(hao)}$
    \STATE \quad \textbf{end if}
    \STATE \quad \textcolor{blue}{$p' \leftarrow$\textsc{AnalyticalUpdate}$(p, \hat{s'}, o, a)$}
    \STATE \quad  \textcolor{blue}{\textbf{append} $(p', \hat{r})$ \textbf{to} $B(hao)$}
    \STATE \quad \textcolor{blue}{\textbf{append} $\mathcal{Z}(o \mid p, a, p')$ \textbf{to} $W(hao)$}
    \STATE \quad \textbf{if} $o \notin C(ha)$ \textbf{then} \quad \textit{ $\triangleright$ new node}
        \STATE \quad \quad $C(ha) \leftarrow C(ha) \cup \{o\}$
\STATE \quad \quad \textcolor{blue}{$total \leftarrow \hat{r} + \gamma$ * \textsc{Rollout}($p'$, $hao$, $d - 1$)}
    \STATE \quad \textbf{else}
        \STATE \quad \quad \textcolor{blue}{$(p',r) \leftarrow \text{select } B(hao)[i] \text{ w.p. } \frac{W(hao)[i]}{\sum_{j=1}^{m} W(hao)[j]}$}
        % \STATE \quad \quad \textcolor{blue}{$r \leftarrow R(\hat{s}, a, s')$}
        \STATE \quad \quad \textcolor{blue}{$total \leftarrow r + \gamma \textsc{Simulate}(p', hao, d - 1)$}
    \STATE \quad \textbf{end if}
    \STATE \quad $N(h) \leftarrow N(h) + 1$
    \STATE \quad $N(ha) \leftarrow N(ha) + 1$
    \STATE \quad $Q(ha) \leftarrow Q(ha) + \frac{total - Q(ha)}{N(ha)}$
    \STATE \quad \textbf{return} $total$
\STATE \textbf{end procedure}
\end{minipage}
\end{algorithmic}}
\end{algorithm*}

\section{Problem Formulation}
\label{problem_formulation}
We consider a robotic search-and-rescue task in an initially unknown indoor environment populated with a set of $N \in \mathbb{N}$ static landmarks, each of which may or may not be associated with a victim. The robot must operate under partial observability, simultaneously localizing itself, estimating landmark geometry, and reasoning about latent semantic information associated with each landmark in order to make sequential decisions that maximize a task-specific objective (i.e., locating all victims in the environment). An overview of the experimental setup is shown in Fig.~\ref{fig:depth_comparison}.
\subsection{Problem Statement}
At each discrete time step $t$, the robot executes an action $a_t \in \mathcal{A}$, transitions to a new state $s_t \in \mathcal{S}$ according to stochastic dynamics, and receives a noisy observation $o_t \in \mathcal{O}$.
Observations are composed of both geometric measurements $z_t$ and semantic measurements $y_t$, such that $o_t = (z_t, y_t)$.
Geometric observations provide information about landmark locations relative to the robot, while semantic observations provide probabilistic evidence regarding latent semantic attributes associated with each landmark. Therefore, the underlying state space contains both continuous geometric variables and discrete semantic variables that must be inferred from noisy observations over time. The robot's objective is to select actions that maximizes the expected cumulative reward by efficiently visiting landmarks with high semantic likelihood.

This decision-making problem can be naturally formulated as a POMDP as detailed below. Importantly, because the state includes not only robot pose but also landmark geometry and latent semantic variable whose dimensionality grows with the number of landmarks, this results in a high-dimensional belief representation that poses significant challenges for sampling-based online POMDP solvers.

\subsection{State Space $\mathcal{S}$}
The agent's state at time $t$ is defined as 
\begin{equation}
s_t = \big( x_t, \{ \theta_{n,t} \}_{n=1}^N, \{ e_{n,t} \}_{n=1}^N, \{ v_{n,t} \}_{n=1}^N \big),
\end{equation}
where $x_t \in \mathbb{R}^{d_x}$ denotes the robot pose (i.e., position and orientation), $\theta_{n,t} \in \mathbb{R}^{d_\theta}$ denotes the geometric location of the $n$-th landmark, $e_{n,t} \in \{0,1\}$ denotes a latent semantic variable indicating the presence or absence of a victim associated with the $n$-th landmark, and $v_{n,t} \in \{0,1\}$ denotes a visited indicator used for bookkeeping and reward evaluation. This structured decomposition will later be exploited in section \ref{rb-representation} to analytically marginalize subsets of the state within the proposed Rao-Blackwellized belief representation.

\subsection{Action Space $\mathcal{A}$}
At each time step, the robot selects an action $a_t = (v_t, \omega_t)$ where $v_t$ denotes the linear velocity and $\omega_t$ denotes the angular velocity. The action space is discretized to include forward, backward, and left and right turning motions. In addition to motion actions, the robot is equipped with a scanning action that does not induce motion but allows the robot to directly observe the latent semantic state of a landmark when the robot is within the sensing range. Successful identification of a victim through this action contributes to the detection reward defined in \ref{reward_definition}.

\subsection {Transition Model $\mathcal{T}$}
The agent dynamics are governed by the robot motion model and deterministic bookkeeping updates while landmark location and latent semantic variables are assumed to remain static. Given state $s_t$ and action $a_t$, the robot pose evolves according to
\begin{equation}
x_{t+1} = h(x_t, a_t) + \delta_t,
\end{equation}
where $h$ denotes a nonlinear unicycle motion model and $\delta_t \sim \mathcal{N}(0, Q_t)$ represents zero-mean Gaussian process noise. While additive Gaussian noise is assumed here for our experimental setup, both the transition and observation models are not required to follow additive noise assumptions. We adopt the notation convention used in \cite{montemerlo_fastslam_2003} in accordancce to the SLAM literature. 

The visited indicators $\{v_{n,t}\}_{n=1}^N$ are updated deterministically when the robot executes a scan action. If the robot is within a predefined radius of landmark $n$, then the corresponding visited flag is set to one. Otherwise, it remains unchanged.

\subsection {Observations and Observation Model $\mathcal{O}$, $\mathcal{Z}$} 
At each time step, the robot receives observations $o_t = (z_t, y_t) =\{(z_{n,t}, y_{n,t})\}_{n\in \mathcal{N}_t}$ consisting of geomeric and semantic measurements associated with a set of visible landmarks $\mathcal{N}_t \subseteq \{1,\dots,N\}$. We assume that data association is known (i.e., each observation is associated with a unique landmark index). While FastSLAM has been extended to handle unknown data association \cite{montemerlo_fastslam_2003}, this assumption allows us to focus on belief representation and its integration within the POMDP framework.

Geometric observations provide noisy landmark positions relative to the robot and are received whenever landmarks are within the sensing range. Specifically, the geometric observation associated with landmark $n \in \mathcal{N}_t$ is modeled as
\begin{equation}
z_{n,t} = g(x_t, \theta_{n,t}) + \epsilon_{n,t},
\end{equation}
where $g$ denotes a nonlinear range-and-bearing measurement function and $\epsilon_{n,t} \sim \mathcal{N}(0, R_t)$ is zero-mean Gaussian noise. Again, additive Gaussian observation noise assumption is not required.

Semantic observations provide information about the latent semantic variable $e_{n,t}$ associated with each landmark. Such observations are also assumed to be available once the robot executes a sensing action within the sensing range. We assume conditional independence across landmarks and sensing modalities. Details of the semantic measurement model and its role in the importance weight computation are provided in Section \ref{semantic_obs_model}.

\subsection {Reward $\mathcal{R}$}
\label{reward_definition}
The reward function is designed to encourage efficient exploration toward potential victim locations while penalizing unnecessary motion and control effort:
{\small
\begin{equation}
\begin{aligned}
R(s_t, a_t, s_{t+1})
=&\;\underbrace{R_{\text{found}}(s_t, a_t, s_{t+1})}_{\text{detection reward}}
+\underbrace{k_p v_t \cos(e_{\psi_t})}_{\text{radial motion reward}} \\
&\hspace{-5.em}-\underbrace{(k_v v_t^2 + k_\omega \omega_t^2)}_{\text{control penalty}}
-\underbrace{k_{\text{lat}} |v_t| \sin^2(e_{\psi_t})}_{\text{lateral motion penalty}}
-\underbrace{k_h (1 - \cos(e_{\psi_t}))}_{\text{heading penalty}},
\end{aligned}
\end{equation}}where $e_{\psi_t}$ denotes the heading error between the robot orientation and the direction toward the nearest, unvisited target landmark with high semantic probability, and $k_p$, $k_v$, $k_\omega$, $k_{\text{lat}}$, and $k_h$ are positive weighting coefficients. The detection reward is defined as 
{\small
\begin{equation}
\begin{aligned}
R_{\text{found}}(s_t, a_t, s_{t+1})\hspace{-.3em} \\
&\hspace{-5.em}=\sum_{n=1}^N
\mathbb{I}\left[
a_t = a_{\text{scan}} \wedge
v_{n,t+1}=1 \wedge
v_{n,t}=0 \wedge
e_{n,t}=1
\right] R_0,
\end{aligned}
\end{equation}}where $R_0$ is a constant positive reward associated with successfully identifying a victim..

\section{Methodology} 
\label{Methodology}
\subsection{Rao-Blackwellized Belief Representation}
\label{rb-representation}
Conditioned on the robot trajectory, landmark states are assumed to be independent, as is standard in FastSLAM-based formulations. The posterior then admits the following factorization
{\small
\begin{equation}
\begin{aligned}
p\left(x^t, \{\theta_{n,t}, e_{n,t}, v_{n,t}\}_{n=1}^N \mid o^t, a^t\right) \\
&\hspace{-10.em}= p(x^t, \{v_{n,t}\}_{n=1}^N \mid o^t, a^t)
\prod_{n=1}^N
p(\theta_{n,t}, e_{n,t} \mid x^t, o^t, a^t)
\end{aligned}
\end{equation}}where $x^t$, $o^t$, and $a^t$ denote the histories of robot trajectory, observations, and actions up to time step $t$, respectively.

The belief is approximated using a set of $M$ Rao-Blackwellized particles. Each particle represents a sampled robot trajectory and deterministic bookkeeping variables, combined with an associated map in which landmark geometry and semantic attributes are maintained analytically. In particular, each particle carries $N$ EKFs for geometric landmark states and $N$ Bernoulli distributions for semantic landmark states. Formally, the $m$-th particle at time $t$ is given by
\begin{equation}
S_t^{[m]}=
\Biggl(
x^{t,[m]},
\Bigl\{\bigl(
\underbrace{\mu_{n,t}^{[m]}, \Sigma_{n,t}^{[m]}}_{\text{EKF}},
\underbrace{\pi_{n,t}^{[m]}}_{\text{Bernoulli }},
v_{n,t}^{[m]}
\bigr)
\Bigr\}_{n=1}^N
\Biggr),
\end{equation}
where $(\mu_{n,t}^{[m]},\Sigma_{n,t}^{[m]})$ denote the sufficient statistics of the Gaussian estimate of landmark geometry maintained by each EKF and $\pi_{n,t}^{[m]}$ denotes the Bernoulli parameter associated with the semantic variable $e_{n,t}$.

\subsection{Proposal Distribution}
The efficiency of particle-based belief updates depends on the proposal distribution used for sampling. We adopt the measurement-informed proposal distribution from FastSLAM~2.0. The key idea is to sample robot poses by incorporating both the control input and the geometric landmakr measurements to reduce particle depletion compared to sampling from the motion model alone \cite{montemerlo_fastslam_2002}. Specifically, the proposal distribution for the robot state at time $t$ is given by
{\small
\begin{equation}
\begin{aligned}
q\left(x_t \mid x_{t-1}^{[m]}, a^t, z^t\right)
&\propto\underbrace{
p\left(x_t \mid x_{t-1}^{[m]}, a_t\right)
}_{\sim \mathcal{N}\!\left(x_t;\, h(x_{t-1}^{[m]}, a_t), Q_t\right)} \ \\
&&\hspace{-16.em}\int
\underbrace{
p\left(z_{n,t} \mid \theta_{n,t}, x_t\right)
}_{\sim \mathcal{N}\!\left(z_{n,t};\, g(\theta_{n,t}, x_t), R_t\right)}
\;
\underbrace{
p\left(\theta_{n,t} \mid x^{t-1,[m]}, z^{t-1}\right)
}_{\sim \mathcal{N}\!\left(\theta_{n,t};\, \mu_{n,t-1}^{[m]}, \Sigma_{n,t-1}^{[m]}\right)}d\theta_{n,t}
\end{aligned}
\end{equation}}Although multiple landmark measurements may be available at time $t$, the proposal is constructed by incorporating landmark measurements
sequentially. For notational clarity, this is illustrated here for a single measurement $z_{n,t}$ following \cite{montemerlo_fastslam_2002}. The above proposal does not admit a closed form solution in general. However, we can approximate the measurement function using a first-order Taylor expansion, yielding a Gaussian proposal distribution. Under this EKF-style approximation, the proposal distribution becomes
\begin{equation}
q(x_t \mid x_{t-1}^{[m]}, a^t, z^t)
=
\mathcal{N}\!\left(x_t;\, \mu_{x_t}^{[m]}, \Sigma_{x_t}^{[m]}\right),
\end{equation}
where $\mu_{x_t}^{[m]}$ and $\Sigma_{x_t}^{[m]}$ are computed as described in \cite{montemerlo_fastslam_2002}.

\subsection{Importance Weight Computation}
Note that semantic observations are not used in the proposal distribution. However, they still must be incorporated in the importance weights computation by marginalizing the continuous landmark geometry, robot's pose, and the discrete semantic variables. Under the assumption of conditional independence across landmarks and sensing modalities, the overall observation likelihood factorizes as 
\begin{equation}
p(o_t \mid s_t)=\prod_{n \in \mathcal{N}_t}p(z_{n,t} \mid x_t, \theta_{n,t}, e_{n,t})\,p(y_{n,t} \mid x_t, \theta_{n,t}, e_{n,t}).
\end{equation}
Therefore, for an observation associated with landmark $n$, the weight update for particle $m$ can be written as
{\small
\begin{equation}
\label{weight_computation}
\begin{aligned}
w_{n,t}^{[m]}
&\propto
p\!\left(z_{n,t}, y_{n,t} \mid x^{t-1,[m]}, a^t, o^{t-1}\right) \\
&=
\sum_{e \in \{0,1\}}
p\left(e_{n,t} = e \mid x^{t-1,[m]}, a^t, o^{t-1}\right) \\[6pt]
&\quad
\iint
p(y_{n,t} \mid  x_t,\theta_{n,t}, e)\,
\underbrace{
p(z_{n,t} \mid x_t, \theta_{n,t})
}_{\sim \mathcal{N}\!\left(z_{n,t};\, g(\theta_{n,t}, x_t), R_t\right)} \\[6pt]
&
\underbrace{
p\!\left(
\theta_{n,t}
\mid
x^{t-1,[m]}, a^{t-1}, z^{t-1}
\right)
}_{\sim \mathcal{N}\!\left(
\theta_{n,t};\, \mu_{n,t-1}^{[m]}, \Sigma_{n,t-1}^{[m]}
\right)}d\theta_{n,t}
\underbrace{
p(x_t \mid x_{t-1}^{[m]}, a_t)
}_{\sim \mathcal{N}\!\left(
x_t;\, h(x_{t-1}^{[m]}, a_t), Q_t
\right)}
\, dx_t .
\end{aligned}
\end{equation}}Following FastSLAM~2.0, the integrals over the robot pose and landmark geometry are again approximated in closed form by linearizing the measurement model $g$, yielding a Gaussian likelihood evaluation. To preserve a fully closed-form importance weight computation, the semantic observation model must be carefully designed, as detailed in the next section \ref{semantic_obs_model}.

\subsection{Semantic Observation Modeling}
\label{semantic_obs_model}
One way to ensure that \eqref{weight_computation} remains analytically tractable is to model the semantic likelihood such that it admits a closed-form expression when marginalized under the Gaussian uncertainty induced by the robot pose and landmark geometry. Specifically, we model semantic detection as a Bernoulli random variable whose success probability is defined as a function of the range defined as $d$:
\begin{equation}
p(y_{n,t} = 1 \mid \theta_{n,t}, x_t, e_{n,t}=1)
= P_D\!\big(d(x_t,\theta_{n,t})\big).
\end{equation}
To preserve analytical tractability, the detection function $P_D(d)$ must be chosen such that marginalization over a Gaussian-distributed range,
\begin{equation}
d \sim \mathcal{N}\!\left(\hat{d}_n^{[m]}, \sigma_{n,d}^{2\,[m]}\right),
\end{equation}
admits a closed-form solution. The mean and variance of the range distribution are again approximated via first-order linearization as
\begin{equation}
\hat{d}_n^{[m]} = f\!\left(\hat{x}_t^{[m]}, \hat{\theta}_{n,t}^{[m]}\right), \quad
\sigma_{n,d}^{2\,[m]} = J_x Q_t J_x^{\top}
+ J_{\theta_n}\, \Sigma_{n,t-1}^{[m]} J_{\theta_n}^{\top},
\end{equation}
where $\hat{x}_t^{[m]}=h(x_{t-1}^{[m]}, a_t)$ and $\hat{\theta}_{n,t}^{[m]}=\mu_{n,t-1}^{[m]}$ denote the predicted robot pose and predicted landmark location, respectively, and $J_x$ and $J_{\theta_n}$ are the Jacobians of $f$ with respect to $x_t$ and $\theta_{n,t}$.

While commonly used link functions such as the logistic function do not yield closed-form expressions when marginalized under Gaussian uncertainty, the probit link yields an exact analytical solution. Using the probit model, we therefore define the detection probabilities as
{\small
\begin{equation}
\label{semantic_model}
\begin{aligned}
\bar{P}_D^{[m]}
&= \mathbb{E}_{d \sim \mathcal{N}(\hat{d}_n^{[m]}, \sigma_{n,d}^{2\,[m]})}
\!\left[ P_D(d) \right] \\
&\hspace{-2.em}= \int \Phi(\alpha_0 + \alpha_1 d)\,
\mathcal{N}\!\left(d;\hat{d}_n^{[m]}, \sigma_{n,d}^{2\,[m]}\right)\, \hspace{-.2em}dd = \Phi\!\left(
\frac{\alpha_0 + \alpha_1 \hat{d}_n^{[m]}}
{\sqrt{1 + \alpha_1^2 \sigma_{n,d}^{2\,[m]}}}
\right),
\end{aligned}
\end{equation}}where $\Phi$ denotes the standard normal cumulative distribution function and $\alpha_0$, $\alpha_1$ are model parameters. A derivation of this closed-form expression is provided in Appendix~\ref{appendix_B}. Under this formulation, the expected detection and similarly expected false-alarm probabilities (i.e. $\bar P_{FA}^{[m]}$ with different model parameters $\beta_0$ and $\beta_1$) can be computed in closed form. This enables the importance weight computation in \eqref{weight_computation} to be performed analytically:
{\small
\begin{equation}
\begin{aligned}
w_{n,t}^{[m]}
&\approx 
\mathcal N\!\big(z_{n,t};\ \hat z_{n,t}^{[m]},\,\Lambda_{n,t}^{[m]}\big)
\Big[
\pi_{n,t}^{[m]}\,
\big(\bar P_D^{[m]}\big)^{y_{n,t}}
\big(1-\bar P_D^{[m]}\big)^{1-y_{n,t}} \\
&\qquad +
\big(1-\pi_{n,t}^{[m]}\big)\,
\big(\bar P_{FA}^{[m]}\big)^{y_{n,t}}
\big(1-\bar P_{FA}^{[m]}\big)^{1-y_{n,t}}
\Big],
\end{aligned}
\end{equation}}where $\hat z_{n,t}^{[m]} =g\left(\hat x_t^{[m]},\, \hat\theta_{n,t}^{[m]}\right)$ is the predicted measurement and the innovation covariance is
\begin{equation}
\Lambda_{n,t}^{[m]}
=
G_xQ_tG_x^{\top}+G_\theta\Sigma_{n,t-1}^{[m]}G_\theta^{\top} + R_t.
\end{equation}
Here, the matrices $G_x$ and $G_\theta$ are the Jacobians of $g$ with respect to $x_t$ and $\theta_{n,t}$, respectively \cite{montemerlo_fastslam_2003}. 

\subsection{Belief update}
\label{belief_update}
The update step for geometric landmark states remains the same as in FastSLAM, which is equivalent to the standard EKF measurement update \cite{montemerlo_fastslam_2003}\cite{maybeck_stochastic_1979}. We therefore focus on the update of the semantic belief associated with each landmark.

The latent semantic state is modeled as a Bernoulli random variable with particle-specific belief $\pi_{n,t}^{[m]}$ for landmark $n$. Hence, given a binary semantic observation $y_{n,t}$, the semantic belief is updated analytically via Bayes’ rule as
{\small
\begin{equation}
\pi_{n,t}^{[m]}
=
\left[
1
+
\frac{1-\pi_{n,t-1}^{[m]}}{\pi_{n,t-1}^{[m]}}
\left(\frac{\bar P_{FA}^{[m]}}{\bar P_D^{[m]}}\right)^{y_{n,t}}
\left(\frac{1-\bar P_{FA}^{[m]}}{1-\bar P_D^{[m]}}\right)^{1-y_{n,t}}
\right]^{-1}.
\end{equation}}

\section{Experimental Results}
\label{experimental_results}
\subsection{Belief Quality and Computation}
\begin{figure}[t]
    \centering
    \includegraphics[width=\linewidth]{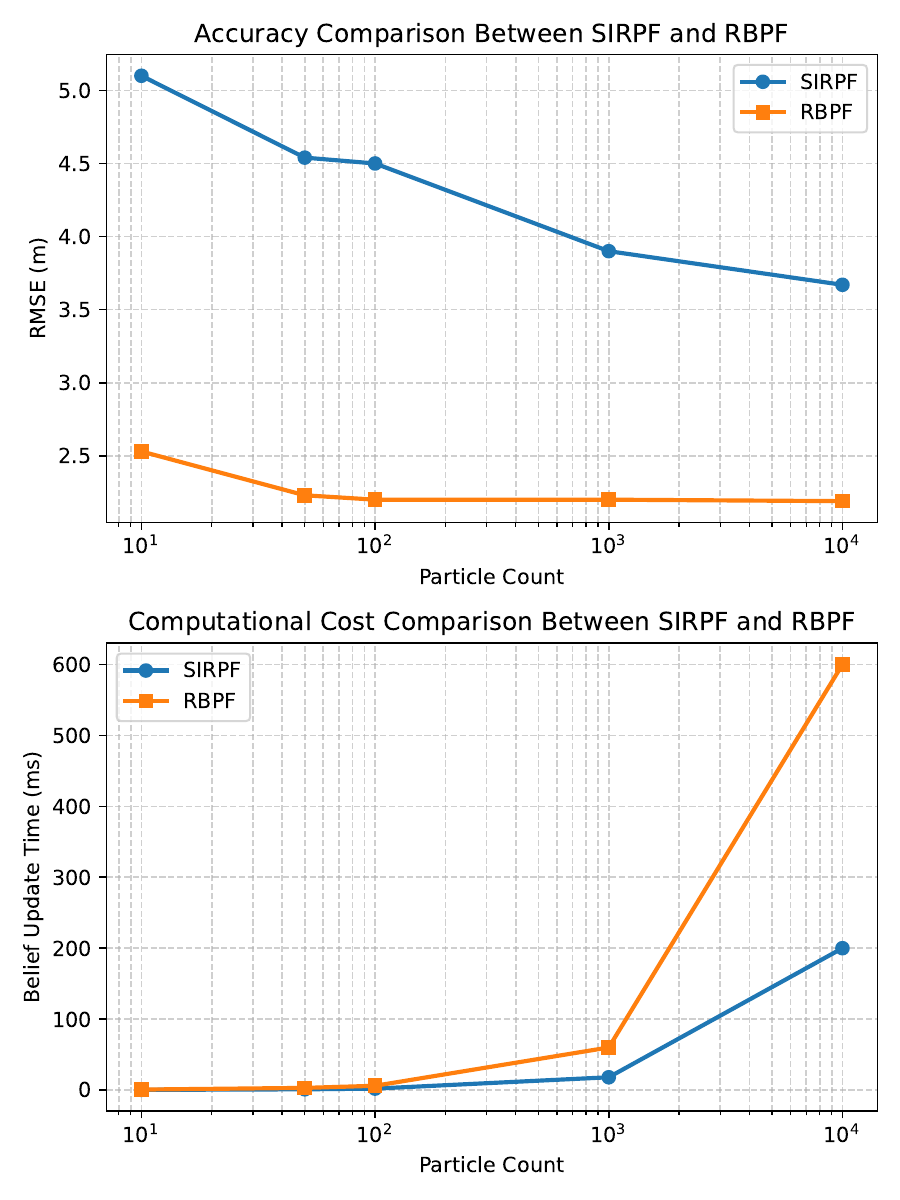}
    \caption{Comparison of estimation accuracy and computational cost between SIRPF and RBPF. RMSE is computed by combining the squared estimation errors of both the robot pose and all observed landmark positions. Computational cost is computed per belief update at each time step.}
    \label{fig:belief_comparison}
\end{figure}
We first compare the accuracy of RBPF and SIRPF by evaluating their root-mean-square error (RMSE). As shown in the top plot of Figure \ref{fig:belief_comparison}, RBPF achieves low RMSE with a relatively small number of particles, after which further increases in particle count yield only marginal improvements. In particular, approximately 50 particles are sufficient for RBPF to near-optimal accuracy. In contrast, SIRPF requires substantially more number of particles to approach comparable accuracy. Even with 10,000 particles, the RMSE of SIRPF remains higher than that of RBPF using as few as 10 particles. This behavior is consistent with prior results in FastSLAM works, where Rao-Blackwellization substantially improves estimation efficiency and accuracy by analytically marginalizing conditionally tractable state subcomponents \cite{montemerlo_fastslam_2003}\cite{montemerlo_fastslam_2002}. 

We next analyze the computational costs of belief updates shown in the bottom plot of Figure \ref{fig:belief_comparison}. For an equal number of particles, RBPF incurs higher per-update computation time than SIRPF due to the additional analytical updates required for each landmark belief. This increase is expected as each Rao-Blackwellized particle maintains an EKF for every landmark geometry as well as a Bernoulli belief for each latent semantic variable. Inevitably, updating these analytical belief components incurs additional per-particle computational overhead compared to SIRPF, which samples all state variables directly. However, because RBPF analytically marginalizes both continuous landmark geometry and discrete semantic variables, it decreases the dimensionality of the sampled state space. As a result, RBPF achieves improved estimation accuracy with significantly fewer particles, reducing the overall computational cost required to maintain a belief of similar quality. Thus, despite higher per-particle costs, RBPF still offers superior efficiency in practice by significantly reducing the number of particles required for accurate belief representation. 

\begin{figure*}[!t]
\centering
\subfloat[\textbf{POMCPOW (SIRPF)}]{%
    \includegraphics[width=2.2in]{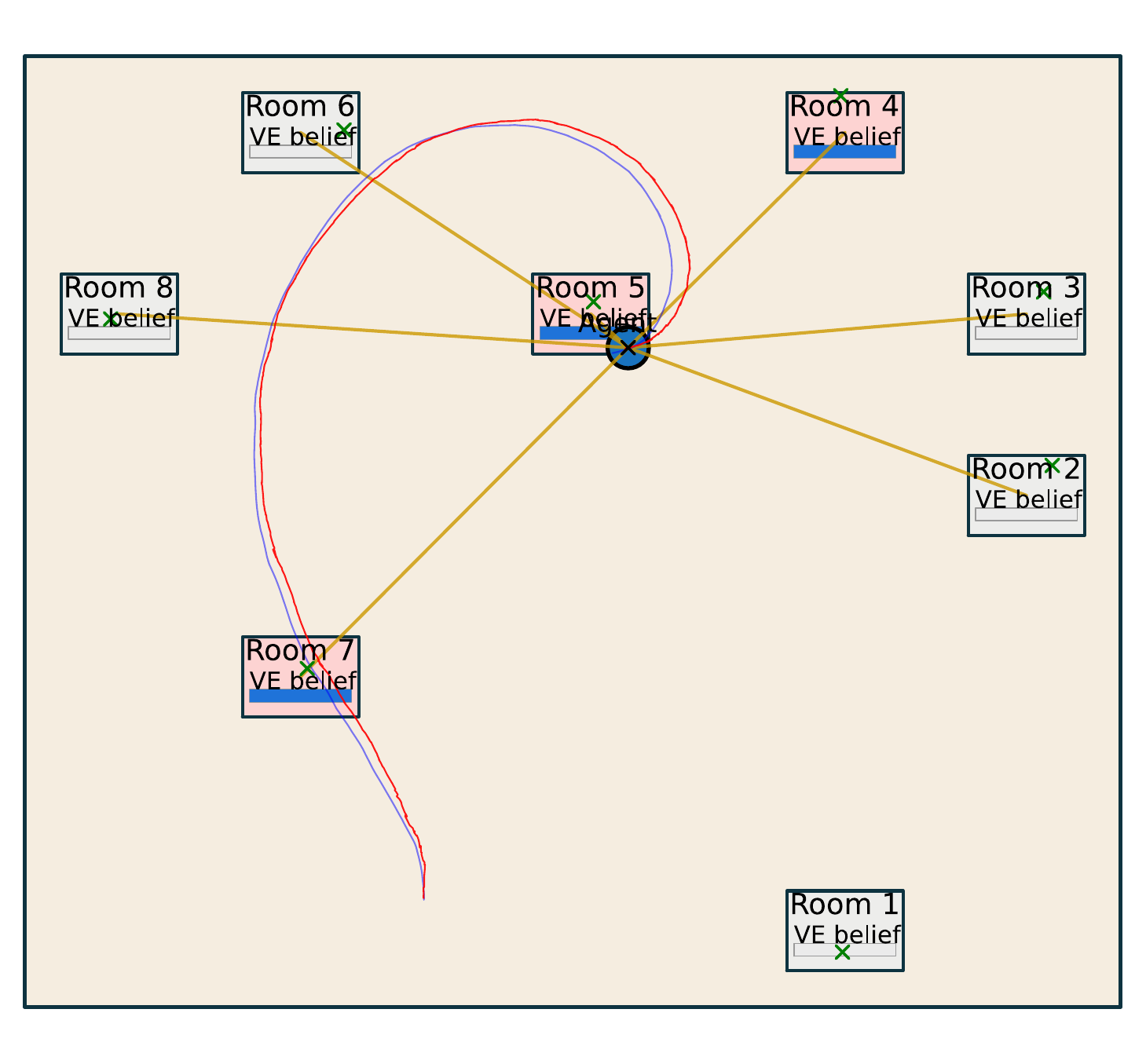}
    \label{fig:depth5}
}
\hfill
\subfloat[\textbf{RB-MC-POMCPOW (RBPF)}]{%
    \includegraphics[width=2.2in]{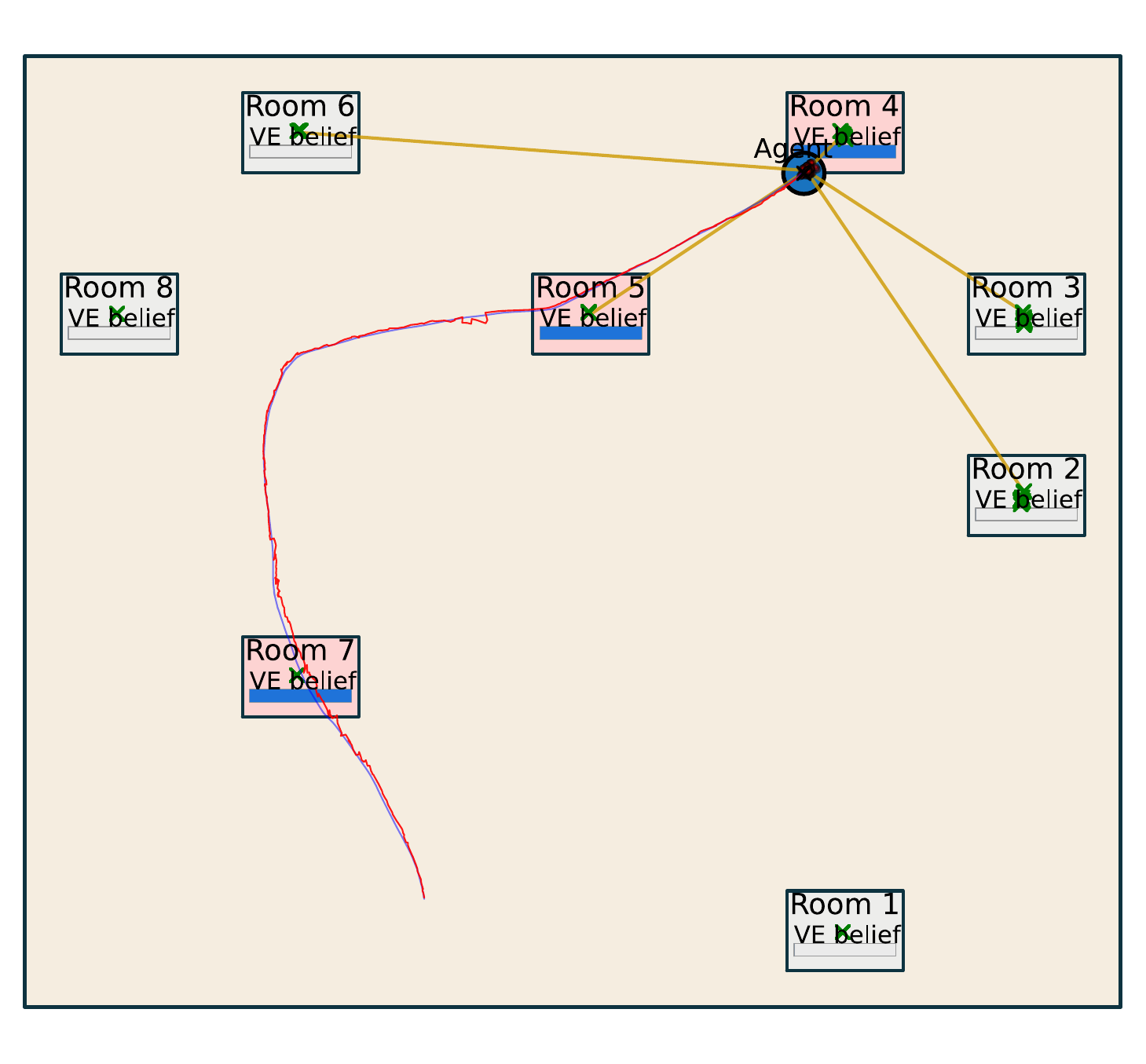}
    \label{fig:depth10}
}
\hfill
\subfloat[\textbf{RB-POMCPOW (RBPF)}]{%
    \includegraphics[width=2.2in]{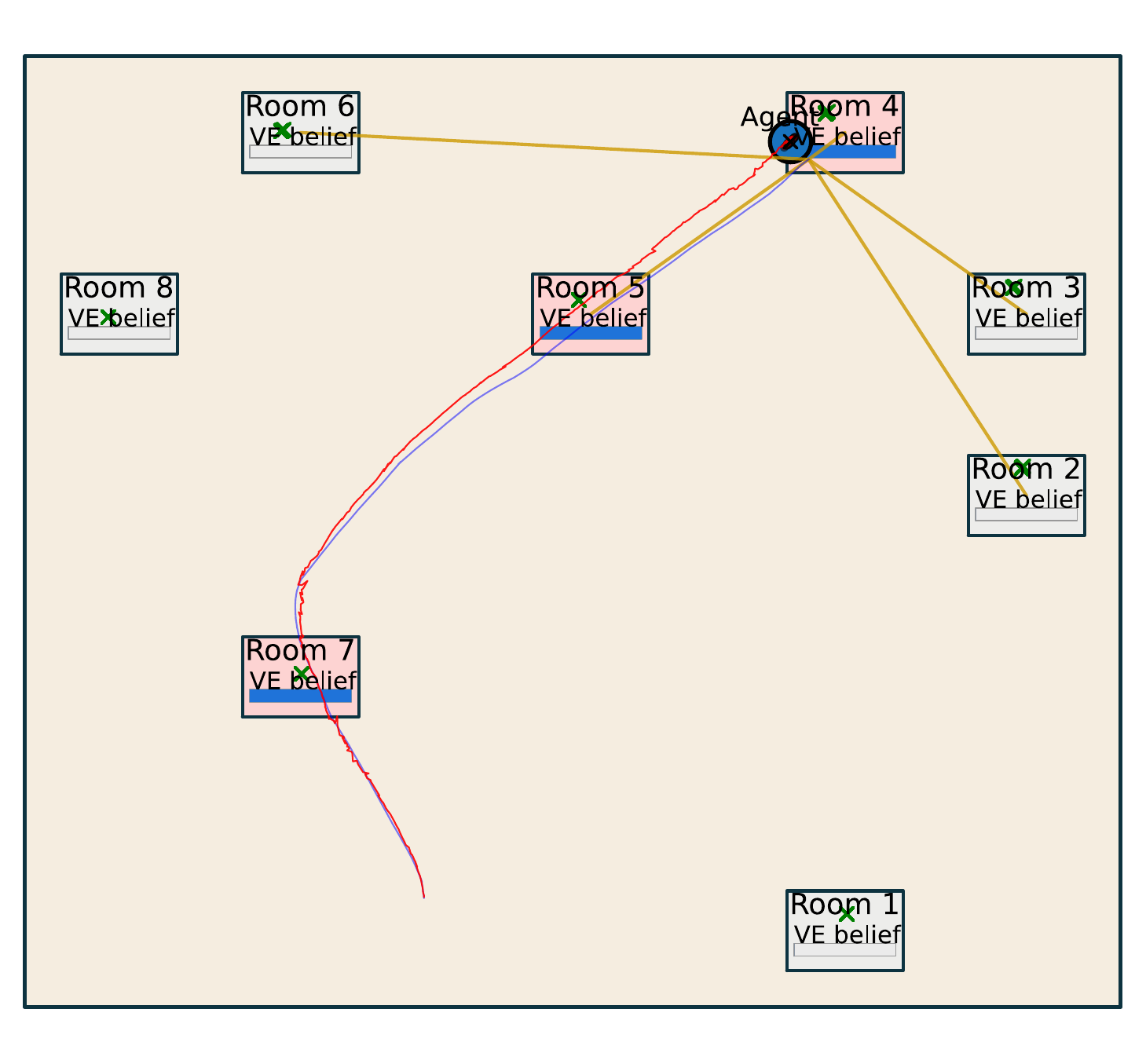}
    \label{fig:depth20}
}
\caption{Effect of planner and belief representation on robot behavior. Representative robot trajectories generated using (a) POMCPOW with a standard SIRPF, (b) RB-MC-POMCPOW with RBPF but performs Monte Carlo–based generative evaluation by drawing a single realization (cf. Monte Carlo scheme in Fig.~\ref{fig:rb_pipeline}), and (c) RB-POMCPOW which instead evaluates expected rewards and observations via analytical integration (cf. Quadrature scheme in Fig.~\ref{fig:rb_pipeline}). Victims are located at the centers of rooms, and the blue bars indicate the corresponding victim-evidence belief which converges toward the true value over time. Red and blue curves denote the estimated and ground-truth robot trajectories, respectively, while green crosses indicate particle or EKF-based estimates of victim locations. All planners are executed with the same maximum time horizon, computational budget, and hyperparameters. In the representative run shown, POMCPOW with SIRPF (a) fails to reach the high-probability victim in Room 4. RB-MC-POMCPOW (b) eventually reaches all of the victim rooms, but the trajectory exhibits unnecessary detours when approaching Room 5. Although these purely sampling-based planners can occasionally produce optimal trajectories, they exhibit higher variability, reflecting sampling-induced variance in value estimation. In contrast, RB-POMCPOW (c) better accounts for uncertainty with analytical integration at each tree node and results in more direct navigation toward rooms with high victim belief.}
\label{fig:depth_comparison}
\end{figure*}

\subsection{Planning Performance and Computation}
\label{planning_performance}
We next evaluate how belief representation and generative evaluation influence planning performance under a fixed planning computational budget. Representative trajectories generated by each planner are shown in Fig.~\ref{fig:depth_comparison}. In Fig.~\ref{fig:depth_comparison} (a), POMCPOW with a standard SIRPF belief exhibits large trajectory deviations and fails to reach the high-probability victim in Room 4. In Fig.~\ref{fig:depth_comparison} (b), we replace the belief representation with RBPF while retaining Monte Carlo–based generative evaluation during tree expansion. In this setting, analytically maintained landmark geometry and semantic states are randomly sampled to produce a single realization when simulating predicted observations and rewards, as illustrated by the Monte Carlo scheme in Fig.~\ref{fig:rb_pipeline}. We denote this planner as RB-MC-POMCPOW. Although RB-MC-POMCPOW benefits from reduced belief estimation variance due to Rao-Blackwellization, the use of stochastic realizations of analytically tractable state components during planning introduces additional planning uncertainty. As a result, the planner eventually locates all victim rooms but exhibits unnecessary detours when approaching Room 5 from Room 7. While both sampling-based planners can occasionally generate near-optimal trajectories given sufficient simulations with increased planning computational budget, their performance remains inconsistent due to sampling-induced variance in predicted rewards and observations during tree expansion.

In contrast, RB-POMCPOW replaces these stochastic realizations with deterministic expectation over analytically tractable state components during generative evaluation. This corresponds to the quadrature scheme in Fig.~\ref{fig:rb_pipeline}. This reduces the variance of predicted outcomes within the planning tree and results in more consistent value estimation, yielding more direct navigation toward rooms with high victim belief, as shown in Fig.~\ref{fig:depth_comparison} (c).

\begin{figure}[t]
    \centering
    \includegraphics[width=\linewidth]{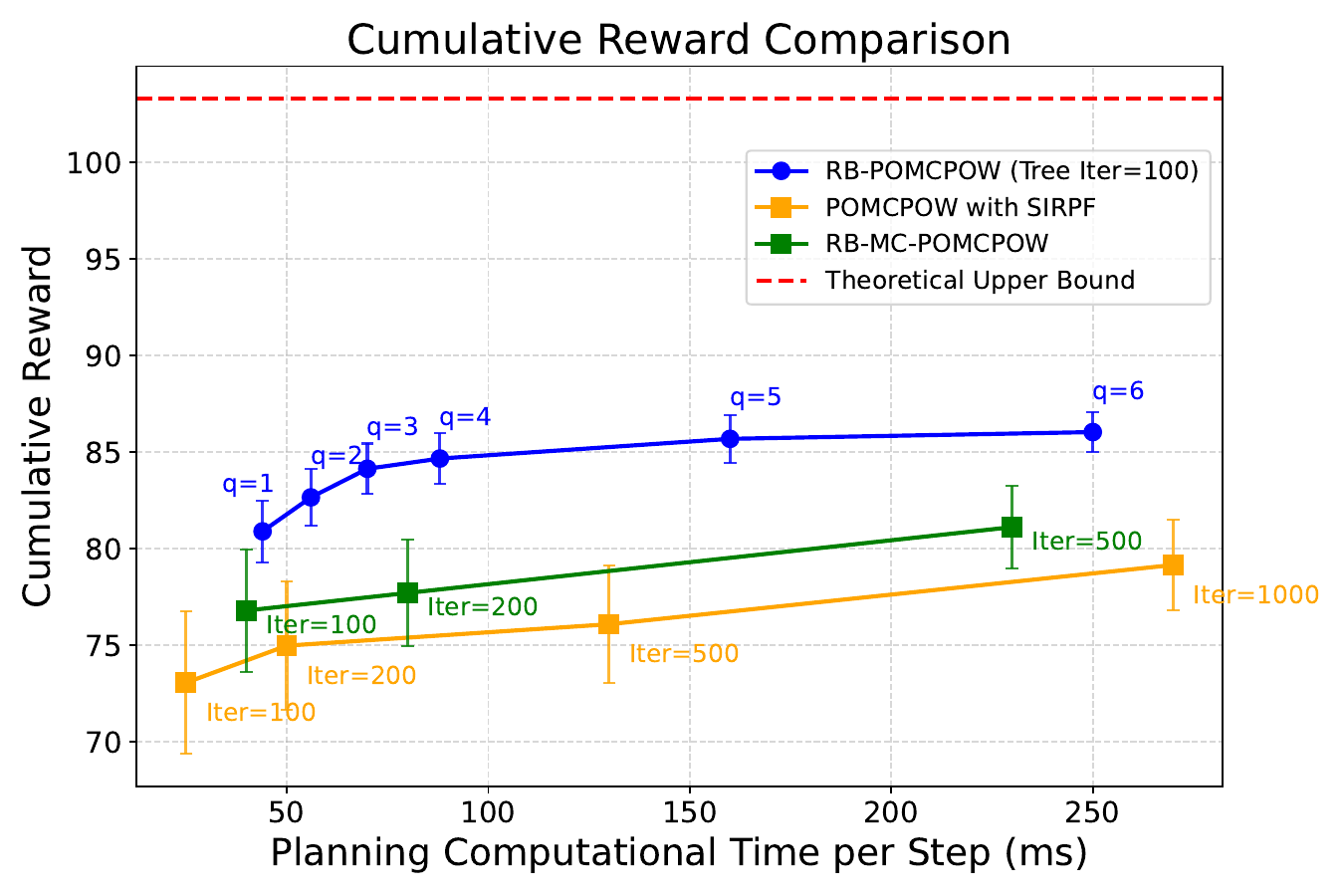}
    \caption{Planning performance versus computational cost. Cumulative reward as a function of per-step planning time for POMCPOW, RB-MC-POMCPOW and RB-POMCPOW. For RB-POMCPOW, results are shown for different sparse grid levels (denoted as $q$) with fixed tree iterations. The dashed red line indicates a theoretical upper bound on cumulative reward obtained by assuming perfect semantic information and optimal trajectory path; therefore, it represents a loose performance reference that cannot be achieved. RB-POMCPOW consistently achieves higher reward than both POMCPOW and RB-MC-POMCPOW under comparable computational budgets.}
    \label{fig:planning_comparison}
\end{figure}

To quantify this effect, Figure \ref{fig:planning_comparison} shows cumulative reward as a function of per-step planning computation time for the baseline POMCPOW, RB-MC-POMCPOW, and RB-POMCPOW planners. To ensure comparable belief estimation accuracy, we use 50 and 10,000 particles for RBPF and SIRPF, respectively, as established in the previous section. For RB-POMCPOW, performance is evaluated across varying sparse grid levels with fixed tree iteration counts. For POMCPOW and RB-MC-POMCPOW, we only vary the tree iterations because both rely on Monte Carlo generative evaluation. 

As expected, increasing either the sparse grid level or the number of tree iterations generally improves cumulative reward by producing more accurate value function estimates. However, computational cost grows proportionally with the sparse grid level for RB-POMCPOW, which limits the number of tree iterations that can be performed within a fixed computational budget. Despite this trade-off, RB-POMCPOW consistently outperforms both POMCPOW and RB-MC-POMCPOW under similar computational budgets. For example, even with 1500 tree iterations, POMCPOW achieves cumulative rewards comparable only to RB-POMCPOW using a sparse grid level of $q=1$ with just 100 tree iterations. Similarly, RB-MC-POMCPOW with 500 tree iterations provides only marginal improvement over RB-POMCPOW with $q=1$, indicating that improved belief representation alone is insufficient to achieve comparable planning performance.

These results can be explained by two complementary variance-reduction mechanisms. First, 
RBPF maintains conditionally independent landmark beliefs analytically within each particle, resulting in higher effective sample size compared to SIRPF. This aligns with the Rao-Blackwell theorem, ensuring that each selected Rao-Blackwellized particle provides a high-quality representation of the belief state with reduced variance in belief representation. Second, RB-POMCPOW computes expected observations and rewards during generative evaluation by integrating over analytically tractable state components, further reducing uncertainty in predicted outcomes during tree expansion. Together, these effects reduce both belief and planning variance, allowing value function estimates to converge with much fewer tree iterations than purely sampling-based planners. 

%% Discussion
\section{Conclusion and Future Work}
\label{conclusion}
This work presented a Rao-Blackwellized online planning framework for high-dimensional POMDPs and validated its effectiveness in an indoor robotic search-and-rescue task by integrating it with the FastSLAM~2.0 algorithm. The results highlight an important relationship between belief quality and planning quality in POMDPs. While RBPFs have long been shown to improve estimation accuracy and efficiency in robotic systems, improving belief performance alone is not the only benefit of Rao-Blackwellization in online planning settings. By analytically marginalizing tractable state components and propagating their uncertainty through deterministic integration, the RB-POMDP framework reduces variance not only in belief estimation but also in value estimation during planning. This reduction in planning variance translates directly into improved computational efficiency, suggesting that Rao-Blackwellized online planning is not merely beneficial but becomes essential for computationally tractable decision-making in high-dimensional problems. 

However, the effectiveness of Rao–Blackwellized planning depends fundamentally on whether expectations of rewards and observations can be computed tractably with respect to the marginalized state components during generative evaluation. In this work, landmark geometry and semantic attributes are represented using Gaussian and Bernoulli distributions, respectively, both of which admit closed-form belief updates and efficient expectation computation during planning.

In contrast, extending the RB-POMDP framework to occupancy grid-based SLAM such as GMapping \cite{grisetti_improving_2005}\cite{grisetti_improved_2007} presents additional challenges. Occupancy grid maps represent each cell using independent Bernoulli random variables, resulting in a very high-dimensional discrete belief space. Although these cell-wise beliefs are analytically tractable, the ray-casting observation model depends jointly on multiple cells along each sensing beam. Therefore, computing the expected rewards and observations during generative evaluation requires marginalization over the joint occupancy configuration of all cells intersecting the ray, resulting in an exponential number of realizations.  

While naive Monte Carlo sampling from independent cell-wise occupancy distributions is computationally feasible, it fails to enforce spatial coherence in the instantiated map realizations and can produce spatially inconsistent checkerboard-like occupancy patterns. Planning over such sampled map can cause the agent to exhibit hallucinated behaviors during tree expansion.

Recent work on generative observation models for range sensing, including learned LiDAR simulators, offers a promising direction for addressing this limitation by producing realistic map realizations \cite{avidan_learning_2022}\cite{zyrianov_lidardm_2025}\cite{caccia_deep_2019-1}\cite{xiong_learning_2023}. Integrating such generative sensing models (e.g. \cite{deglurkar_compositional_2021}) into the Rao-Blackwellized online planning represents an important avenue for future work to enable integrating POMDPs in richer SLAM settings beyond feature-based representations.

%%\section{Acknowledgement}
% Draper (need to hide)
% TODO XXX: Add Zach's funding 

% \section{Tables}
% Note that, for IEEE-style tables, the
%  $\backslash${\tt{caption}} command should come BEFORE the table. Table captions use title case. Articles (a, an, the), coordinating conjunctions (and, but, for, or, nor), and most short prepositions are lowercase unless they are the first or last word. Table text will default to $\backslash${\tt{footnotesize}} as
%  the IEEE normally uses this smaller font for tables.
%  The $\backslash${\tt{label}} must come after $\backslash${\tt{caption}} as always.
 
% \begin{table}[!t]
% \caption{An Example of a Table\label{tab:table1}}
% \centering
% \begin{tabular}{|c||c|}
% \hline
% One & Two\\
% \hline
% Three & Four\\
% \hline
% \end{tabular}
% \end{table}

% {\appendix[Proof of the Zonklar Equations]
% Use $\backslash${\tt{appendix}} if you have a single appendix:
% Do not use $\backslash${\tt{section}} anymore after $\backslash${\tt{appendix}}, only $\backslash${\tt{section*}}.
% If you have multiple appendixes use $\backslash${\tt{appendices}} then use $\backslash${\tt{section}} to start each appendix.
% You must declare a $\backslash${\tt{section}} before using any $\backslash${\tt{subsection}} or using $\backslash${\tt{label}} ($\backslash${\tt{appendices}} by itself
%  starts a section numbered zero.)}

{\appendices
\section*{Appendix A. RB-POMCP}
\label{RB-POMCP}
The following pseudocode outlines the RB-POMCP algorithm. The code is adapted from \cite{silver_monte-carlo_2010} with the modified code written in blue. The \textsc{Rollout} function is the same as the one defined in Algorithm \ref{alg:RB-POMCPOW}.
\begin{algorithm}
\caption{RB-POMCP}
\begin{algorithmic}[1]
\STATE \textbf{procedure} \textsc{Simulate}($p$,$h$, $depth$)
\STATE \quad \textbf{if} $\gamma^{depth} < \epsilon$ \textbf{then}
\STATE \quad \quad \textbf{return} 0
\STATE \quad \textbf{end if}
\STATE \quad \textbf{if} $hao \notin T$ \textbf{then}
\STATE \quad \quad \textbf{for all} $a \in A$ \textbf{do}
\STATE \quad \quad \quad $T(ha) \leftarrow (N_{\text{init}}(ha), V_{\text{init}}(ha), \emptyset)$
\STATE \quad \quad \textbf{end for}
\STATE \quad \quad \textcolor{blue}{\textbf{return} \textsc{Rollout}$(p, h, depth)$}
\STATE \quad \textbf{end if}
\STATE \quad $a \leftarrow \arg\max_{b} \left[ V(hb) + c\sqrt{\frac{\log N(h)}{N(hb)}} \right]$
\STATE \quad \textcolor{blue}{$(\hat{s'}, \hat{o}, \hat{r}) \leftarrow \mathbb{E}[G(p, a)]$} \hfill \textcolor{black}{$\triangleright$ Quadrature Techniques}
\STATE \quad \textcolor{blue}{$p' \leftarrow$ \textsc{AnalyticalUpdate}$(p, \hat{s'}, \hat{o}, a)$}
\STATE \quad \textcolor{blue}{$R \leftarrow \hat{r} + \gamma$ * \textsc{Simulate}($p'$, $ha\hat{o}$, $depth + 1$)}
\STATE \quad $N(h) \leftarrow N(h) + 1$
\STATE \quad $N(ha) \leftarrow N(ha) + 1$
\STATE \quad $V(ha) \leftarrow V(ha) + \frac{R - V(ha)}{N(ha)}$
\STATE \quad \textbf{return} $R$
\STATE \textbf{end procedure}
\end{algorithmic}
\end{algorithm}
}
\section*{Appendix B. Closed-Form Marginalization of Probit Detection Models under Gaussian Uncertainty}
\label{appendix_B}
This appendix derives the closed-form expression
\begin{equation}
\label{eq:probit_closed_form}
\int \Phi(\alpha_0+\alpha_1 d)\,\mathcal{N}(d;\mu,\sigma^2)\,dd
=
\Phi\!\left(\frac{\alpha_0+\alpha_1 \mu}{\sqrt{1+\alpha_1^2\sigma^2}}\right),
\end{equation}
which is used in the main text with $\mu=\hat d_n^{[m]}$ and $\sigma^2=\sigma_{n,d}^{2\,[m]}$ in \eqref{semantic_model}. Let $d \sim \mathcal{N}(\mu,\sigma^2)$ and define the probit detection model
\begin{equation}
P_D(d)=\Phi(\alpha_0+\alpha_1 d).
\end{equation}
We seek the marginalized detection probability:
\begin{equation}
\bar P_D \triangleq \mathbb{E}\!\left[\Phi(\alpha_0+\alpha_1 d)\right]
=
\int \Phi(\alpha_0+\alpha_1 d)\,\mathcal{N}(d;\mu,\sigma^2)\,dd.
\end{equation}
We introduce an independent standard normal random variable $u \sim \mathcal{N}(0,1)$ with $u \perp d$. Using the identity $\Phi(x)=\mathbb{P}(u \le x)$, we can rewrite
\begin{equation}
\bar P_D
=
\mathbb{E}_d\!\left[\mathbb{P}\!\left(u \le \alpha_0+\alpha_1 d \mid d\right)\right]
=
\mathbb{P}\!\left(u \le \alpha_0+\alpha_1 d\right).
\end{equation}
Equivalently,
\begin{equation}
\bar P_D
=
\mathbb{P}\!\left(u - \alpha_1 d \le \alpha_0\right).
\end{equation}
Now define the random variable
\begin{equation}
w \triangleq u - \alpha_1 d.
\end{equation}
Since $u$ and $d$ are Gaussian and independent, $w$ is also Gaussian with mean and variance
\begin{align}
\mathbb{E}[w] &= \mathbb{E}[u] - \alpha_1 \mathbb{E}[d] = 0 - \alpha_1 \mu = -\alpha_1 \mu, \\
\mathrm{Var}(w) &= \mathrm{Var}(u) + \alpha_1^2 \mathrm{Var}(d)
= 1 + \alpha_1^2 \sigma^2.
\end{align}
Therefore,
\begin{equation}
w \sim \mathcal{N}\!\left(-\alpha_1 \mu,\; 1+\alpha_1^2\sigma^2\right).
\end{equation}
Hence,
\begin{align}
\bar P_D
&=
\mathbb{P}(w \le \alpha_0)
=
\Phi\!\left(\frac{\alpha_0 - \mathbb{E}[w]}{\sqrt{\mathrm{Var}(w)}}\right) \\
&=
\Phi\!\left(
\frac{\alpha_0 - (-\alpha_1 \mu)}{\sqrt{1+\alpha_1^2\sigma^2}}
\right)
=
\Phi\!\left(
\frac{\alpha_0 + \alpha_1 \mu}{\sqrt{1+\alpha_1^2\sigma^2}}
\right),
\end{align}
which proves \eqref{eq:probit_closed_form}.

\bibliographystyle{IEEEtran}
\bibliography{RB-POMDP}

% \end{IEEEbiography}
% \end{IEEEbiography}
% \end{IEEEbiography}
% \end{IEEEbiography}
% \vfill

\end{document}